\documentclass[11pt]{article}

\usepackage{graphicx}
\usepackage{amsmath}
\usepackage{booktabs}
\usepackage{subcaption}
\usepackage[margin=1.2in]{geometry}
\usepackage{epstopdf}
\usepackage{bm}

\newcommand{\xbf}{{\bm x}}
\newcommand{\argmax}{\mathrm{argmax}}

\begin{document}

\title{A Probabilistic Machine Learning Approach to Detect Industrial Plant Faults}\vspace{0.1in}
\author{Wei Xiao\vspace{0.1in}\\
\small{\textit{SAS Institute Inc, Cary, NC, 27513, USA}}\\
\footnotesize{Wei.Xiao@sas.com}}
\date{}

\maketitle
\baselineskip=20pt

\begin{quotation}
\noindent {\it Abstract:}
Fault detection in industrial plants is a hot research area as more and more sensor data are being collected throughout the industrial process. Automatic data-driven approaches are widely needed and seen as a promising area of investment. This paper proposes an effective machine learning algorithm to predict industrial plant faults based on classification methods such as penalized logistic regression, random forest and gradient boosted tree. A fault's start time and end time are predicted sequentially in two steps by formulating the original prediction problems as classification problems. The algorithms described in this paper won first place in the Prognostics and Health Management Society 2015 Data Challenge. \par

\vspace{9pt}
\par
\end{quotation}\par

\newpage
\section{Introduction}

Fault detection in industrial plants is a hot research topic as more and more sensor data are being collected throughout the industrial process, and standard systems based on univariate statistical process control lack power in these more complex systems.  Early detection of faults can help to avoid system shut-down and component failure or even catastrophes \cite{korbicz2012fault}. 

Many machine learning algorithms used in pattern classification are now being utilized in fault detection. Dimension reduction techniques, such as principal component analysis, partial least squares, and Fisher's discriminant analysis have been applied to detect faults in chemical processes \cite{chiang2000fault,chiang2004fault,yin2012comparison}.  Support vector machine and artificial neural networks are also widely used methods for fault detection; they have been applied to gearbox failure detection \cite{samanta2004gear} and chemical process fault diagnosis \cite{wang2005fault}.  K-Nearest Neighbor and fuzzy-logic are two other powerful methods that have been used to detect faults in semiconductor manufacturing processes \cite{he2007fault} and mechanical systems \cite{korbicz2012fault}. Tree based algorithms such as random forest and gradient boosted tree are useful machine learning algorithms in situations where one expects nonlinear and interactive effects between covariates. They have been applied to fault detection in aircraft systems \cite{lee2014fault}.

This year's Prognostics and Health Management  (PHM) Society data challenge focused on plant fault detection. We try many of the above machine learning techniques and ultimately use a combination of several in our final detection strategy described herein. The rest of the paper is organized as follows. Section 2 discusses the data challenge problem. Section 3 introduces the relative methodologies and our algorithm. Finally, Section 4 concludes the paper and discusses future work.

\section{Problem Statement}
The objective of this year's challenge is to design an algorithm to predict plant faults. Correct prediction involves predicting the type of fault (one of five), as well as the start and end time of each fault, within one hour. 

Three datasets are given, training, test, and validation; they contain information on 33, 15, and 15 plants, respectively. For each plant three files are provided: plant-\#a.csv, plant-\#b.csv, plant-\#c.csv, where \# is the plant id. File (a) contains time series readings of 4 sensors ($S1$-$S4$) and 4 reference signals ($R1$-$R4$) from each plant component. The number of components ($Nm$) varies by plant; data on $Sj$ and $Rj$ for $i$th component are denoted $mi\_Sj$ and $mi\_Rj$, respectively, where $i\in\{1,\ldots,Nm\}$ and $j\in\{1,2,3,4\}$.  File (b) contains time series data for cumulative energy consumed ($E1$) and instantaneous power ($E2$) from a fixed number of zones within a given plant. Each plant zone covers one or more of the plant components and the number of zones ($Nn$) varies by plant. The notation $ni\_Ej$ is used to represent the reading of $Ej$ for the $i$th zone. File (c) contains plant fault events, each characterized by a start time, an end time, and a failure type. Data are given on 6 different fault types ($F1-F6$), but only faults 1-5 are to be predicted. The training dataset has complete fault event data, and is used to train the model. The test dataset has complete fault event data for the first half of the sample, but approximately 50\% of the events in the second half of the data have been randomly removed. The boundary between the first and second half of the data is given, and referred to as the boundary time. Our goal is to predict the deleted fault events. The validation dataset is similar in structure to the test dataset. 

Each team participating in the contest is permitted to submit their predictions of the missing faults in the test data (fault type, and start and end time) once each week to assess their prediction performance and use the score as feedback to improve their model. The final team rank is determined by the score of a submission of predictions based on the validation dataset. The data can be download from NASA Ames Prognostics Data Repository \cite{phm}.   

\subsection{Data Description and Preprocessing}
We began our analysis by first studying the data to garner any information that would be useful in predicting the faults.  Not only do the number of both zones and components vary by plant, but the proportion of each fault type ($PFi$) varies quite dramatically. To illustrate, Table \ref{summary1} summarizes the data for the first five plants in both the training and test datasets. Note that $F3$ (fault type 3) never occurs in plants 2, 3, 5 and 42, and $F5$ never occurs in plants 2, 3, 5, 41, 42 and 46. We also notice that $S3$, $R1$, $R2$, $R3$, $R4$ appear to be categorical variables, and $S1$, $S2$, $S4$ appear to be continuous variables. The number of unique levels of all categorical variables for the same sample plants are summarized in Table \ref{summary2}. Given the above differences across plants and variables, we built a separate model for each plant. 

\begin{table}[ht]  
	\begin{center}  
		\caption{Summary statistics of faults by plant. $Nm$: number of components; $Nn$: number of zones; $PF1$-$PF6$: proportion of each fault type. (Plants in training set: 1, 2, 3, 4, 5; plants in test set: 41, 42, 43, 45, 46)}
		\label{summary1}
		\begin{tabular}{lrrrrrrrr}
			\toprule
			\textbf{Plant} &  \textbf{Nm} &  \textbf{Nn} &    \textbf{PF1} &   \textbf{PF2} &    \textbf{PF3} &    \textbf{PF4} &    \textbf{PF5} &    \textbf{PF6} \\
			\midrule
			1 &   6 &   3 &  0.25 &  0.18 &  0.12 &  0.04 &  0.11 &  0.29 \\
			2 &  13 &   2 &  0.46 &  0.03 &  0.00 &  0.02 &  0.00 &  0.49 \\
			3 &  10 &   2 &  0.17 &  0.01 &  0.00 &  0.02 &  0.00 &  0.80 \\
			4 &   8 &   4 &  0.30 &  0.16 &  0.02 &  0.02 &  0.02 &  0.48 \\
			5 &   3 &   2 &  0.27 &  0.07 &  0.00 &  0.07 &  0.00 &  0.58 \\
			\midrule
			41 &   5 &   2 &  0.29 &  0.10 &  0.02 &  0.05 &  0.00 &  0.54 \\
			42 &  10 &   3 &  0.39 &  0.15 &  0.00 &  0.07 &  0.00 &  0.40 \\
			43 &   6 &   2 &  0.08 &  0.16 &  0.02 &  0.22 &  0.11 &  0.41 \\
			45 &   7 &   2 &  0.17 &  0.47 &  0.05 &  0.06 &  0.03 &  0.22 \\
			46 &   5 &   2 &  0.19 &  0.21 &  0.14 &  0.05 &  0.00 &  0.41 \\
			\bottomrule
		\end{tabular}
	\end{center}
\end{table}

\begin{table}[ht]  
	\begin{center}  
		\caption{Counts of unique levels of all categorical variables. (Plants in training set: 1, 2, 3, 4, 5; plants in test set: 41, 42, 43, 45, 46)}
		\label{summary2}
		\begin{tabular}{lrrrrr}
			\toprule
			\textbf{Plant} &              \textbf{S3} &  \textbf{R1} &  \textbf{R2} &  \textbf{R3} &  \textbf{R4} \\
			\midrule
			1 &  12 &  38 &   6 &   8 &   3 \\
			2 &  11 &  26 &   6 &   6 &   3 \\
			3 &  12 &  30 &   7 &   8 &   3 \\
			4 &  12 &  34 &   7 &   7 &   3 \\
			5 &  12 &  12 &   7 &   6 &   3 \\
			\midrule
			41 &  12 &  33 &   4 &   6 &   3 \\
			42 &   8 &  38 &   5 &   7 &   3 \\
			43 &  12 &  23 &   3 &   4 &   3 \\
			45 &  12 &  23 &   4 &   5 &   3 \\
			46 &  12 &  40 &   4 &   5 &   3 \\
			\bottomrule
		\end{tabular}
	\end{center}
\end{table}

The sampling interval for the data provided was theoretically 15 minutes, however some logging delays resulted in irregular intervals. To preprocess the data, we rounded all timestamps to obtain regular 15-minute gaps, and then combined all three files. We define new variables $TTF\_Fk$, $k=1,\ldots,6$, to represent time to failure of fault type $k$. A negative value, $-i$, means the next fault is $i$ intervals in the future (1 interval is 15 minutes), and a positive value, $i$, means the current fault started $i$ intervals ago and has not yet ended. We define $E3$ as the first order difference of $E1$, i.e., $E3(t) = E1(t) - E1(t-1)$. $E3$ measures the energy consumed in the most recent 15 minutes, which similar as $E_2$ is a way to measure instantaneous power. We also define $start\_Fk$, $k=1,\ldots,6$, as a binary indicator of whether any type $k$ fault starts within one hour of the corresponding timestamp, and define $end\_Fk$, $k=1,\ldots,6$, as the binary indicator of whether any type $k$ fault ends within one hour of the corresponding timestamp. Occasionally observations of covariates on some timestamps are missing. Forward imputation was applied to all covariates to impute these missing values, except for $TTF\_Fk$, $start\_Fk$ and $end\_Fk$, which were imputed with values -999, 0 and 0, respectively. To illustrate these preprocessing steps, a small proportion of plant 1's data are shown in Table \ref{sample_data_prepared}. The imputation simplifies the analysis, and from the authors' observation, it has little influence on the modeling results. 

\begin{table}[ht]   
	\begin{center}  
		\caption{A sample of data from plant 1 after preprocessing.}
		\label{sample_data_prepared}
		\begin{tabular}{lrrrrr}
			\toprule
			\textbf{Timestamp} &  \textbf{m1\_R1} &  \textbf{m1\_S1} &  \textbf{TTF\_F1} &  \textbf{start\_F1} &  \textbf{end\_F1} \\
			\midrule
			2009-09-04 09:00:00 &    739 &    763 &  -7 &     0 &      0 \\
			2009-09-04 09:15:00 &    739 &    763 &  -6 &     0 &      0 \\
			2009-09-04 09:30:00 &    739 &    759 &  -5 &     0 &      0 \\
			2009-09-04 09:45:00 &    700 &    711 &  -4 &     1 &      0 \\
			2009-09-04 10:00:00 &    700 &    711 &  -3 &     1 &      0 \\
			2009-09-04 10:15:00 &    700 &    712 &  -2 &     1 &      0 \\
			2009-09-04 10:30:00 &    700 &    720 &  -1 &     1 &      1 \\
			2009-09-04 10:45:00 &    700 &    714 &   0 &     1 &      1 \\
			2009-09-04 11:00:00 &    700 &    716 &   1 &     1 &      1 \\
			2009-09-04 11:15:00 &    700 &    711 &   2 &     1 &      1 \\
			2009-09-04 11:30:00 &    700 &    720 & -41 &     1 &      1 \\
			2009-09-04 11:45:00 &    700 &    716 & -40 &     1 &      1 \\
			2009-09-04 12:00:00 &    700 &    712 & -39 &     0 &      1 \\
			2009-09-04 12:15:00 &    700 &    711 & -38 &     0 &      1 \\
			2009-09-04 12:30:00 &    700 &    716 & -37 &     0 &      1 \\
			2009-09-04 12:45:00 &    700 &    718 & -36 &     0 &      0 \\
			\bottomrule
		\end{tabular}
	\end{center}
\end{table}

There are segments of time where all covariates are missing and fault type 6 is happening. We assumed the plant must be in some type of maintenance mode during these periods, and we excluded these observations in the following analysis.

\subsection{Visualization}
Visualization was key to our understanding of the data.

\begin{figure}[b]
	\begin{subfigure}{1\textwidth}
		\centering
		\includegraphics[scale=0.45]{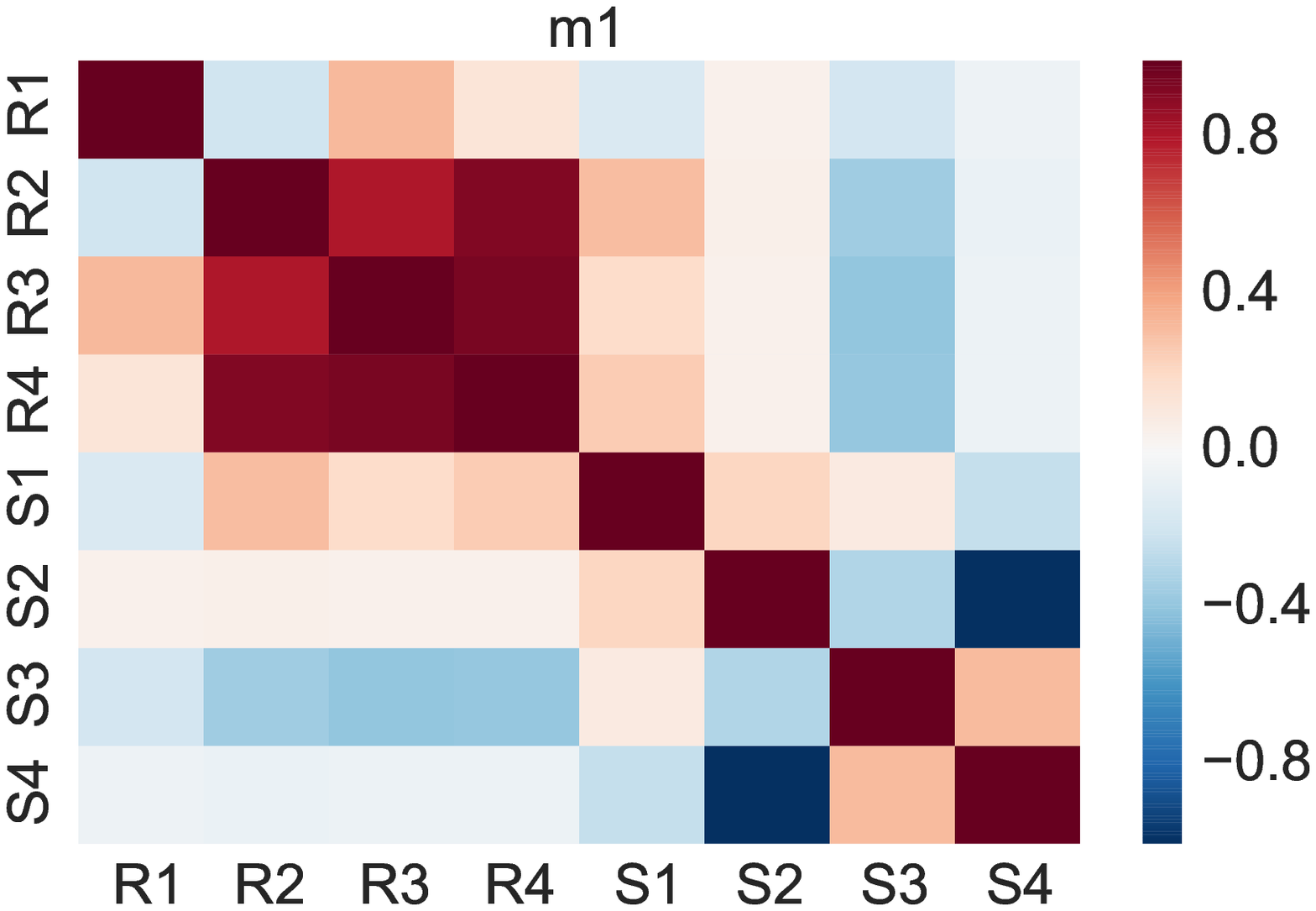}
		\caption{Component 1}
	\end{subfigure}
	\\
	\begin{subfigure}{1\textwidth}
	\centering	
		\includegraphics[scale=0.45]{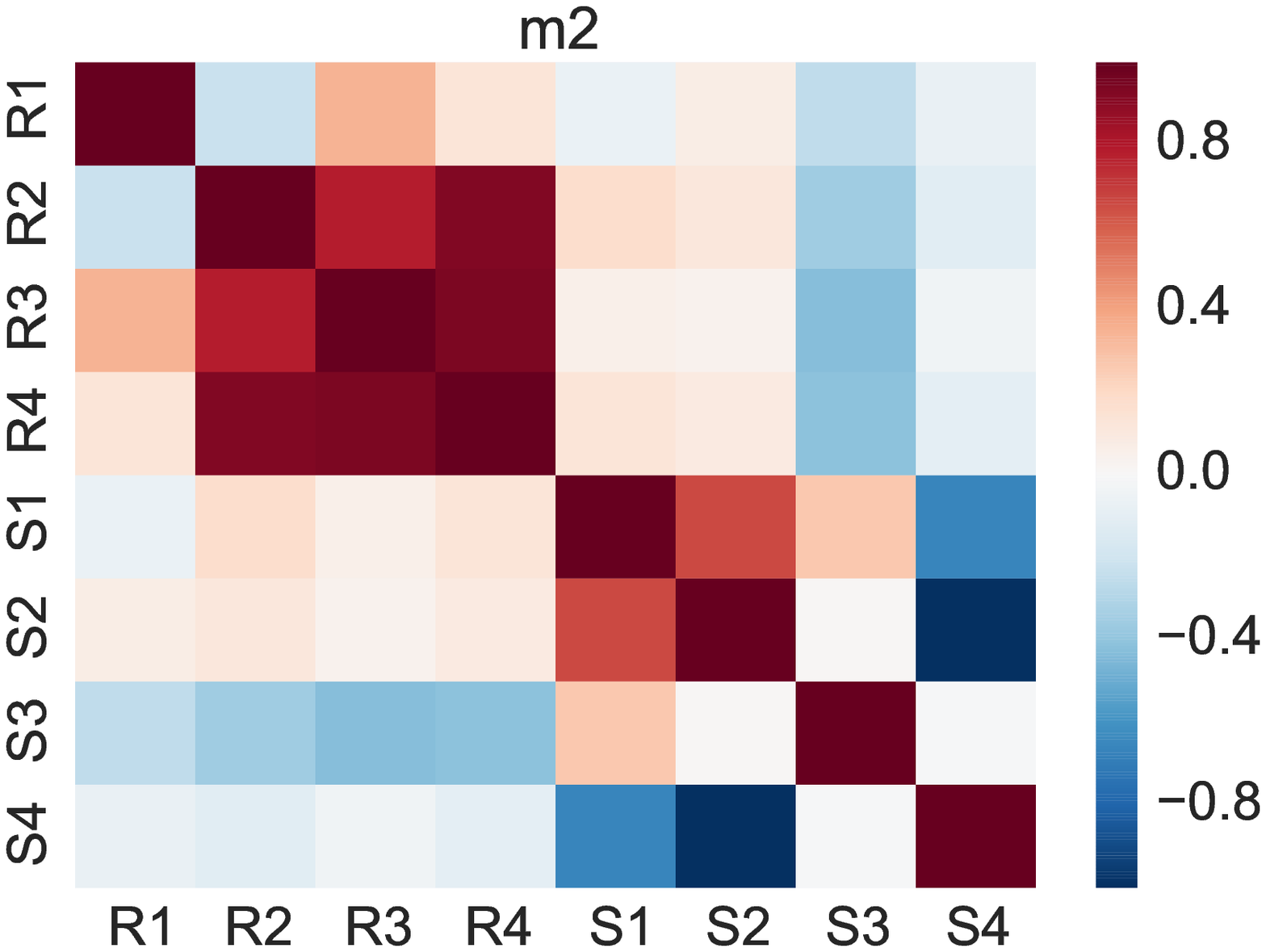}
		\caption{Component 2}
	\end{subfigure}
	\caption{Correlation heatmap for the first two components of plant 1.}
	\label{fig:correlation_heatmap_plant1}
\end{figure}

First, we observe that $R2$, $R3$ and $R4$ are highly positively correlated, and $S2$ and $S4$ are highly negatively correlated, across all components in all plants. To illustrate this finding, Figure \ref{fig:correlation_heatmap_plant1} shows the correlation heatmap of plant 1 for the first two components, where each cell represents the Pearson correlation between two features. Pearson correlation is calculated as
\begin{equation*}
	r=\frac{\sum_{i=1}^n(x_i-\bar{x})(y_i-\bar{y})}{\sqrt{\sum_{i=1}^{n}(x_i-\bar{x})^2}\sqrt{\sum_{i=1}^{n}(y_i-\bar{y})^2}}.
\end{equation*}

\begin{figure}[b]
	\centering
	\includegraphics[scale=0.45]{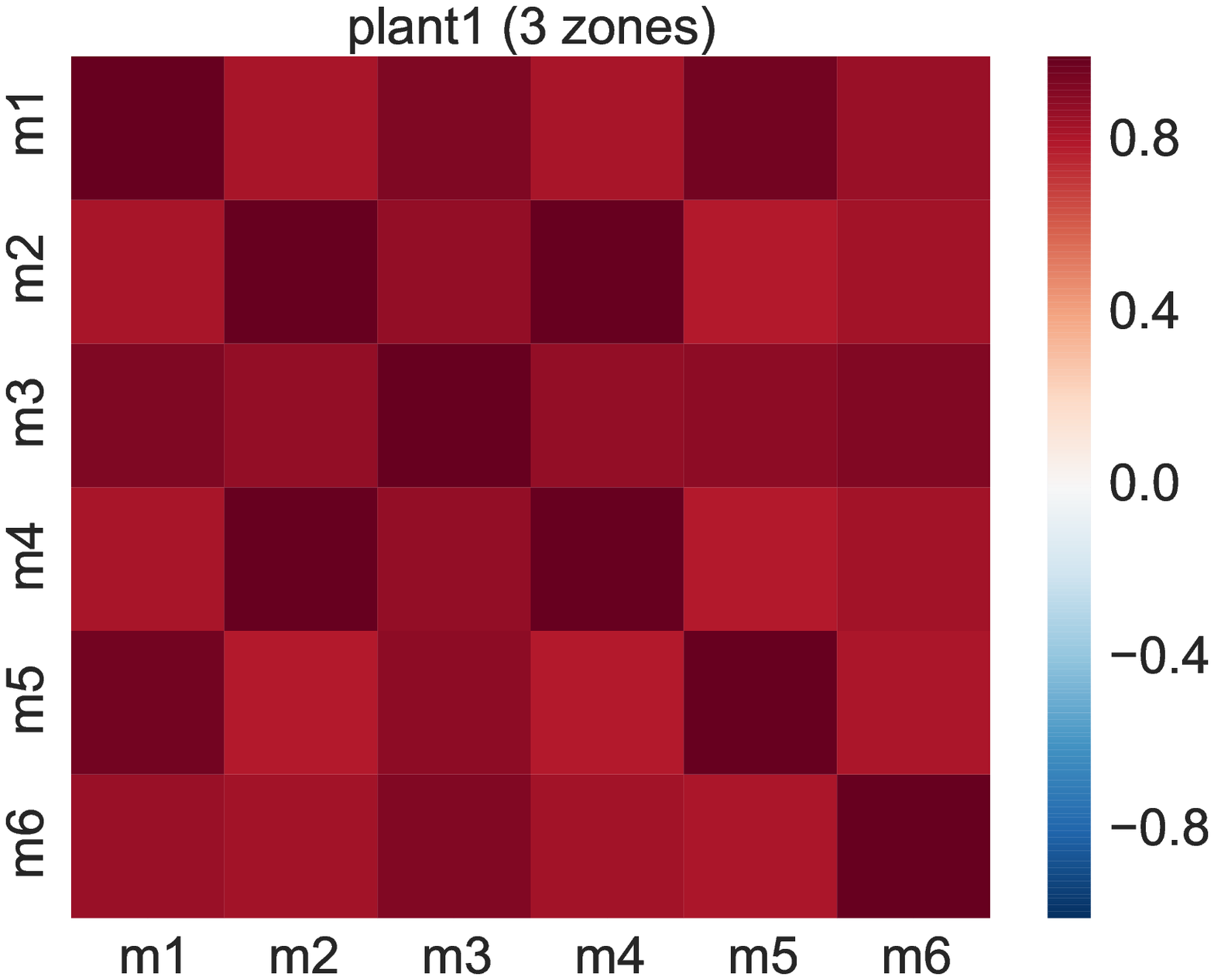}
	\caption{Correlation heatmap of $R4$ across all components in plant 1.}
	\label{fig:correlation_heatmap_zone_plant1}
\end{figure}

Second, by observing the correlation heatmap of either $mi\_R2$, $mi\_R3$, or $mi\_R4$, across all components for a given plant, one can identify which components are in the same zone; components in the same zone are highly correlated. For example, Figure \ref{fig:correlation_heatmap_zone_plant1} shows the correlation heatmap of $mi\_R4$ across all 6 components in plant 1. Based on the heatmap, it seems components 1, 3, 5 of plant 1 belong to one zone, components 2, 4 belong to another zone, and component 6 itself belongs to the third zone. Although one can identify which components are zoned together, the groups of components could not always be linked to a specific zone, so this information was ultimately not utilized in our modeling approach.

\begin{figure}[b]
	\centering
	\includegraphics[scale=0.45]{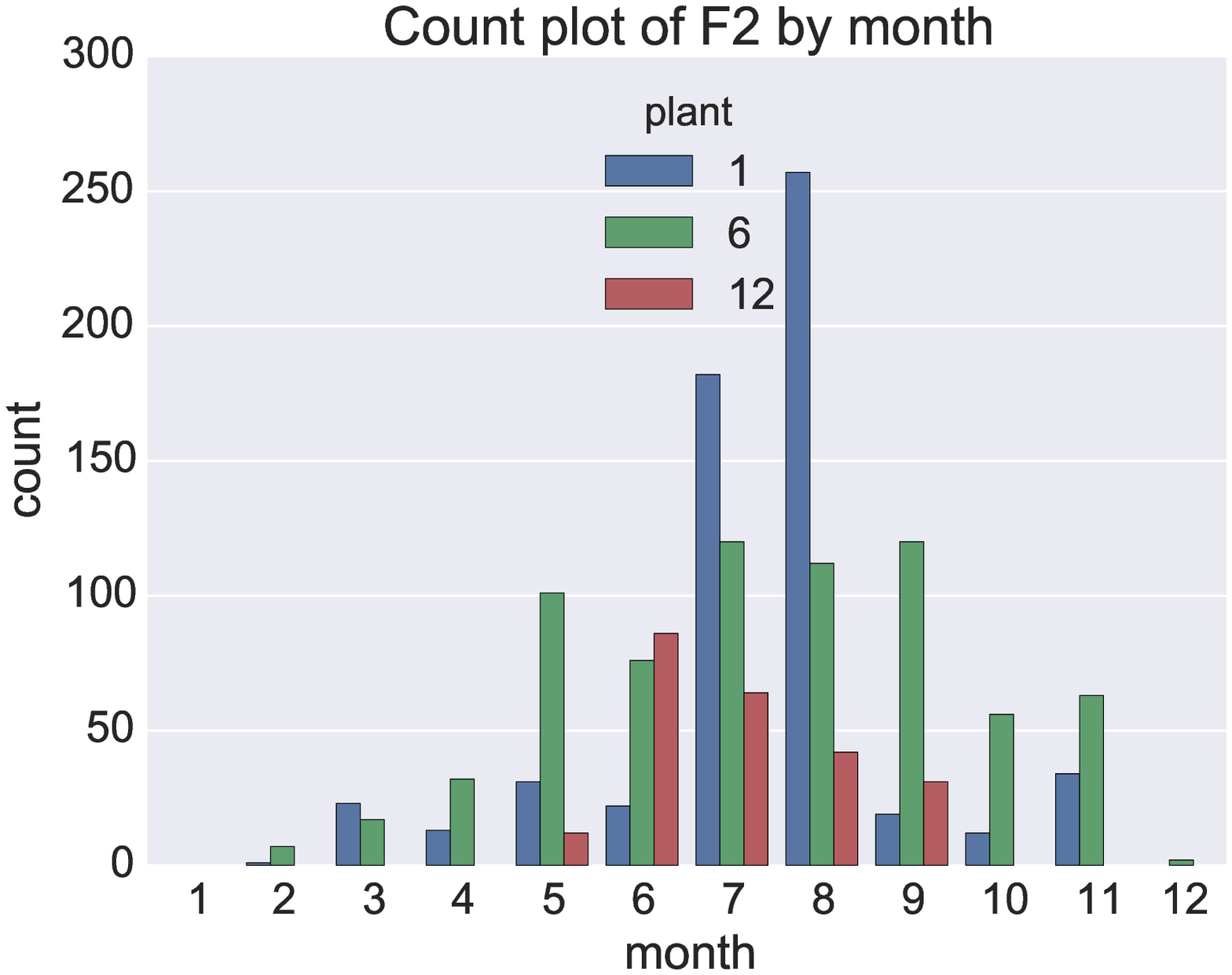}
	\caption{Histogram of fault 2 start times by month (January = 1) for plants 1, 6 and 12.}
	\label{fig:countplot_f2_month}
\end{figure}

\begin{figure}[b]
	\centering
	\includegraphics[scale=0.45]{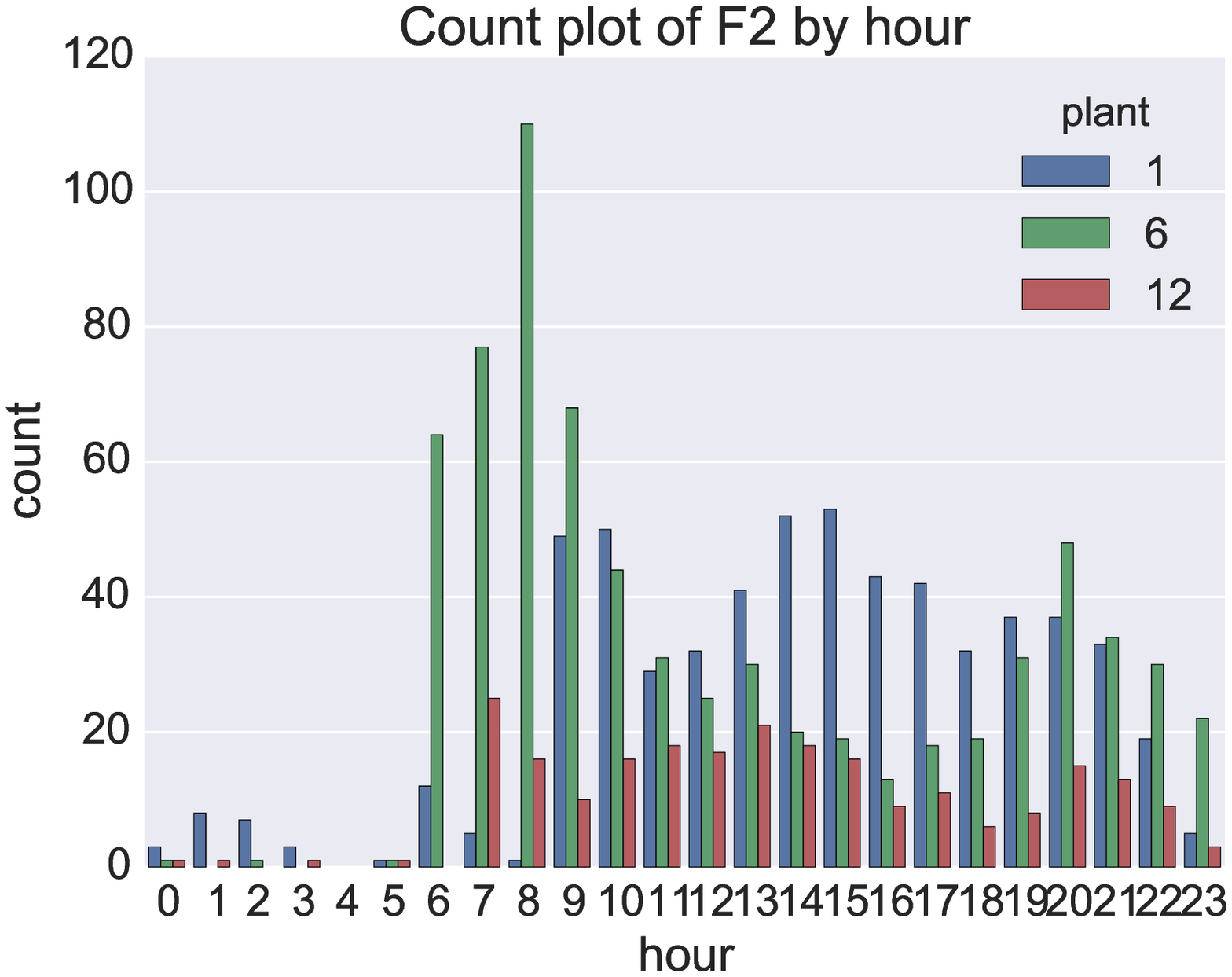}
	\caption{Histogram of fault 2 start times by hour for plants 1, 6 and 12.}
	\label{fig:countplot_f2_hour}
\end{figure}

We also find that month and hour are important categorical variables to predict the faults. Count plots of $F2$ by month and hour are shown in Figure \ref{fig:countplot_f2_month} and \ref{fig:countplot_f2_hour} to illustrate this point. $F2$ starts most frequently between May and November and between 6 o'clock and 23 o'clock. But its distribution varies across plants. 

\begin{figure}[h]
	\centering
	\includegraphics[scale=0.45]{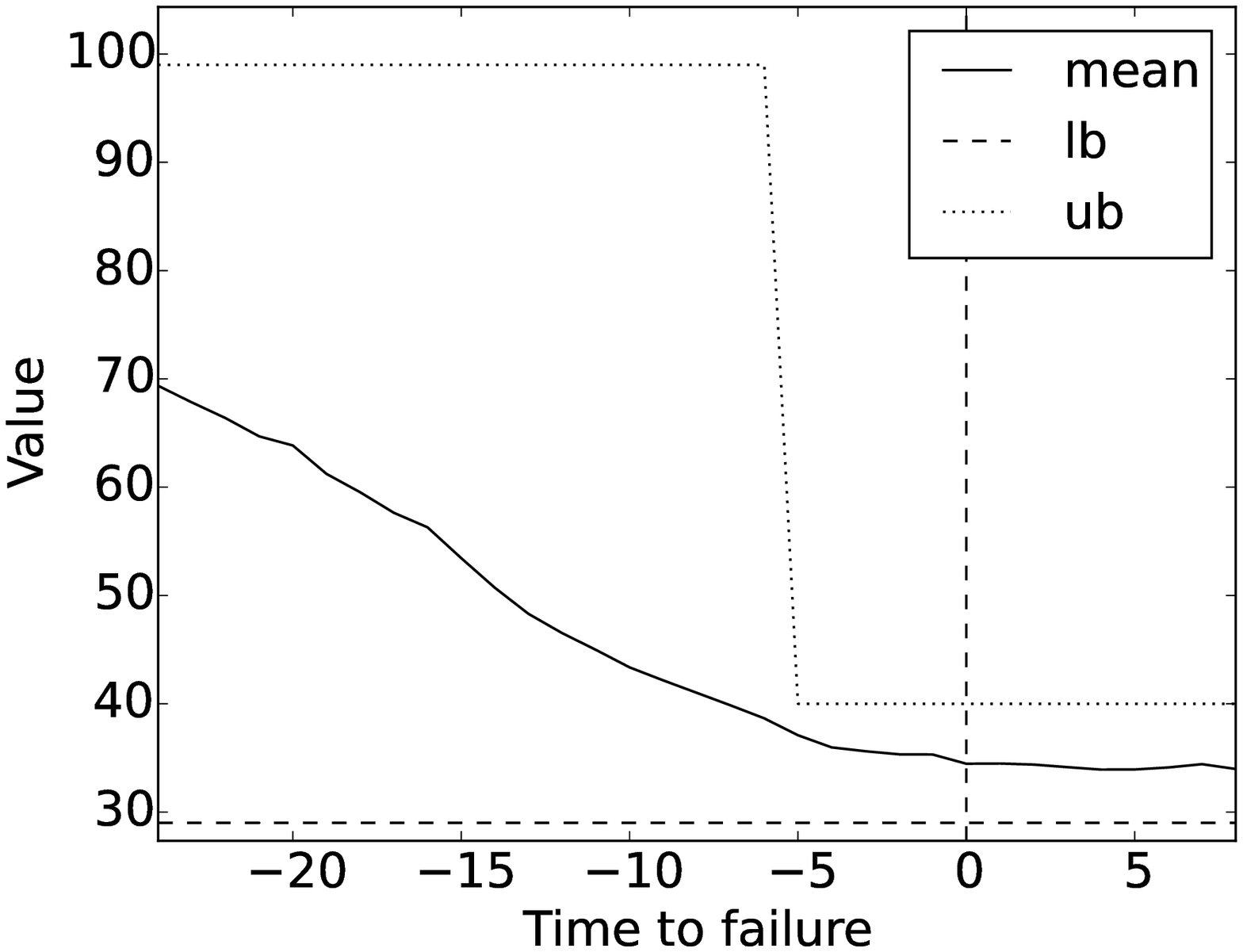}
	\caption{Plot of $m2\_R2$ against time to failure of F1.  lb and ub represents 95\% lower bound and upper bound respectively.}
	\label{fig:m2_R2_plant13}
\end{figure}

Lastly, we observe that, before a fault happens, sensor readings are often increasing or decreasing. These unique patterns can be utilized to predict the start time of the fault. See Figure \ref{fig:m2_R2_plant13} for an example, where the mean value of $m2\_R2$ and its corresponding 95\% confidence bands are plotted against time to failure of $F1$.

\section{Methodology}

\begin{figure}[b]
	\centering
	\includegraphics[scale=0.45]{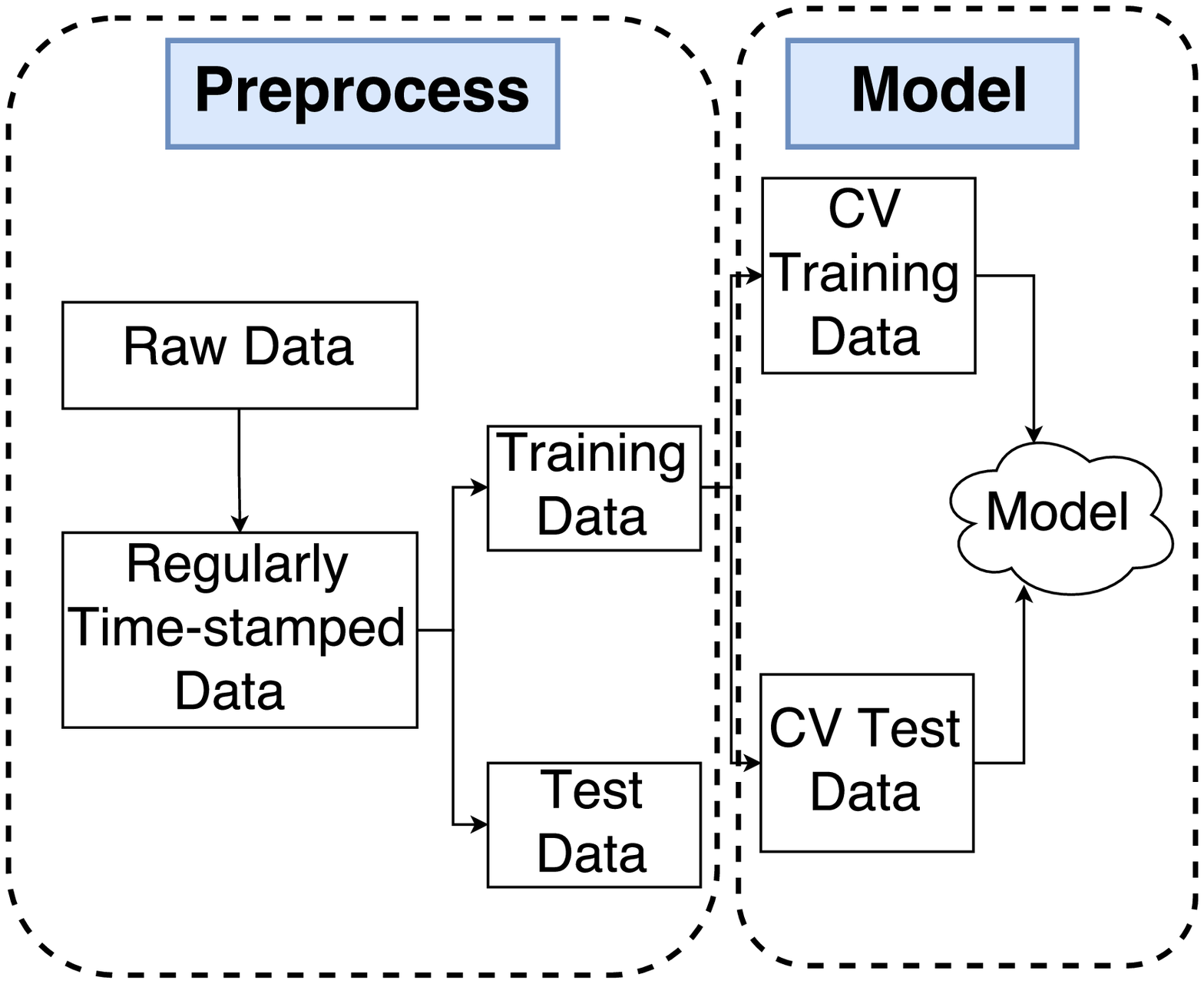}
	\caption{Overall flowchart.}
	\label{fig:overall_flow}
\end{figure}

In this section we introduce our approach and the related methodologies utilized for the PHM competition. The overall approach consists of two parts: preprocessing and modeling. Figure \ref{fig:overall_flow} provides an overview of the process implemented. Details of the data preprocessing have been discussed in Section 2.1. After preprocessing, we divide the training data into two parts: cross validation training data and cross validation test data. Mimicking the test dataset and the validation dataset, the cross validation training data has complete fault event data for the first half of the sample, and 50\% randomly selected events in the second half. The cross validation test data contains the 50\% deleted events in the second half. Our basic approach is to try various models using the cross validation training data and then evaluate their performance based on their ability to forecast faults in the cross validation test data. The winning model is then applied to the test data and the subsequent predictions submitted to PHM for assessment. Here we are not learning the exact model with cross validation training and test data, as we fit a plant specific model to each plant. However, we learn things such as which classifier to use, which threshold value to apply, etc. Please refer to Section 3.3-3.5 for more details.

\begin{figure}[b]
	\centering
	\includegraphics[scale=0.45]{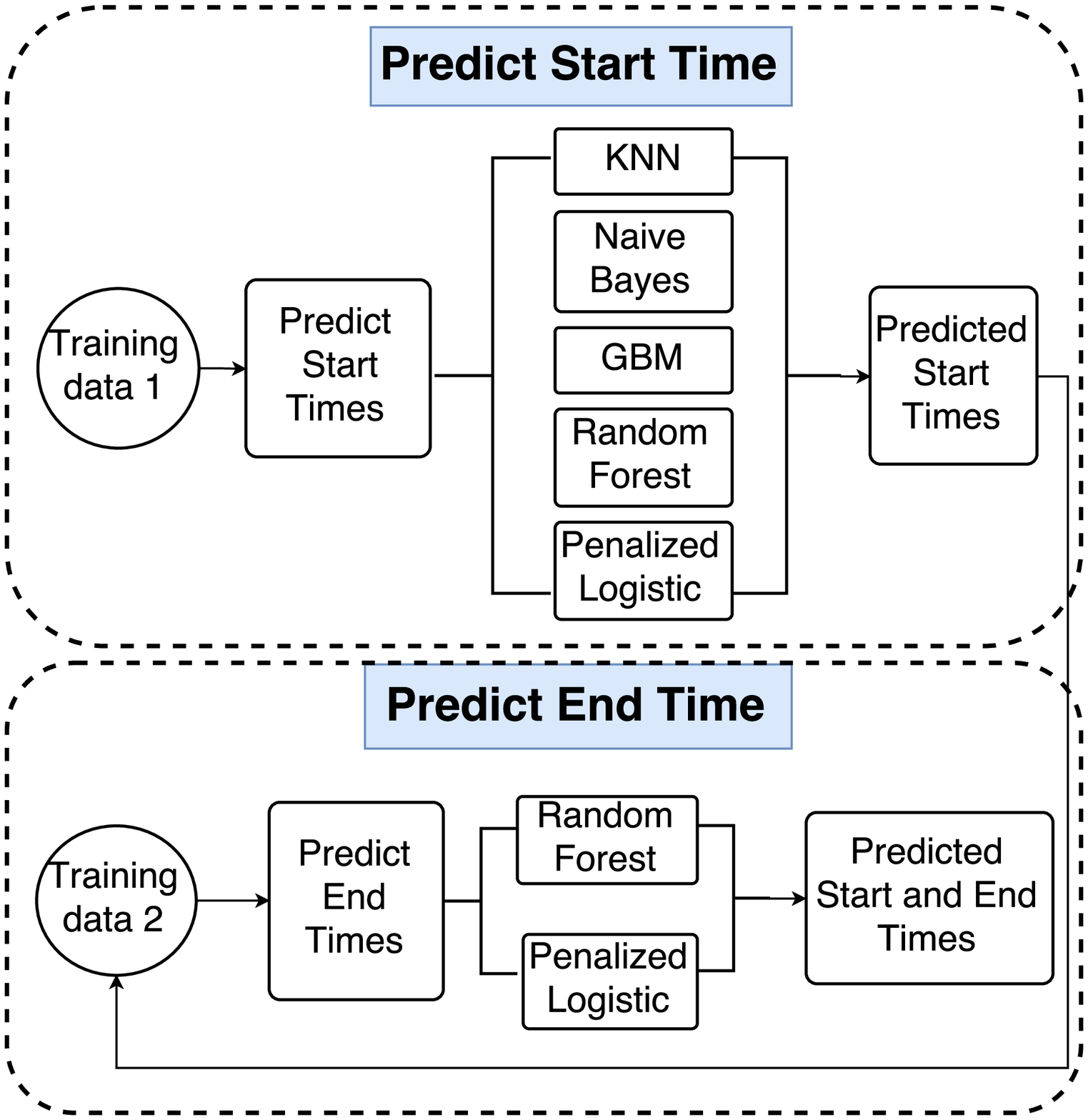}
	\caption{Modeling flowchart.}
	\label{fig:modeling_flow}
\end{figure} 

There are two steps to the modeling process: predict fault start times and then, given these start times, predict fault end time. A detailed flowchart of the modeling process is shown in Figure \ref{fig:modeling_flow}. The modeling procedure outlined is implemented for each fault type, plant by plant. Given a fault type and plant, $F1$ in plant 5 for example, we translate the prediction problem into a classification problem ($start\_F1$=1 vs $start\_F1$=0). From the classification model we estimate the probability that $F1$ starts during each time interval. We derive the set of predicted fault start times, $\Omega_{F1}$, based on these estimated probabilities. For each start time in $\Omega_{F1}$, we then solve another classification problem ($end\_F1$=1 vs $end\_F1$=0) and estimate the probability the $F1$ will end in the next 1 to $t_{\mathrm{max}}$ time intervals, where $t_{\mathrm{max}}$ is an estimated upper bound of fault $F1$'s duration. These estimated probabilities are used to find the fault end time.

Various machine learning algorithms were tried to solve these classification problems: K-nearest neighbors (KNN), naive bayes, gradient boosting machine (GBM), random forest, penalized logisitic regression (with $\ell_2$ penalty), etc. In the final algorithm, we use gradient boosting machine, random forest and penalized logisitic regression. All methods are implemented in SAS or Python Scikit-learn. 

In Section 3.1, we give a quick review of the machine learning algorithms used. Section 3.2 discusses how we evaluate the effectiveness of these different approaches. Section 3.3 describes all the features we use in the model and finally, the specific algorithm details to predict a fault's start time and end time are given in Section 3.4 and 3.5, respectively.  

\subsection{Machine Learning Algorithms}

Data-driven or statistical approaches based solely on historical data are seen as the most cost-effective approach for fault detection in complex systems \cite{aldrich2013unsupervised}. Machine learning is the key to any data-driven algorithm.

Machine learning algorithms can be categorized as either supervised or unsupervised. In supervised learning, the goal is to predict a response $Y$ based on input features $X$. All methods in our algorithm belong to supervised learning.

K-nearest neighbor is an instance-based learning algorithm which has a very simple form but works extremely well on many problems. The algorithm is simple. For a test point $x_0$, we first find $k$ training points that are closest in distance to $x_0$, and then classify using majority vote. K-nearest neighbor can learn very flexible decision boundaries. However, when dealing with high dimensional data, it is likely to suffer from over-fitting and perform poorly due to the curse of dimensionality. 

Naive Bayes \cite{rish2001empirical} is a classification technique based on applying Bayes' theorem. It assumes conditional independence between features given a class $Y=i$. Given a class response $Y$ and a p-dimensional feature $\xbf=(x_1,\ldots,x_p)$, we have
$$\Pr(Y=i|x_1,\ldots,x_p)\propto\Pr(Y=i)\prod_{k=1}^{p}\Pr(x_k|Y=i)$$ based on Bayes' theorem and conditional independence assumption, and Naive Bayes classifies $Y$ as $$\hat{Y}=\argmax_{i}\Pr(Y=i|x_1,\ldots,x_p).$$
Despite its oversimplified and sometimes unrealistic conditional independence assumption, it often outperforms other more sophisticated algorithms. Naive Bayes is widely used in text mining and natural language processing. 

Logistic regression \cite{hosmer2004applied} is widely used in classification problems. However, when the number of input features is large, it performs poorly due to overfitting.  Penalized logisitic regression avoids the overfitting problems of logistic regression by imposing a penalty on large fluctuations in the estimated parameters. In this paper, we use a penalized logistic regression with $\ell_2$ penalty. Besides avoiding overfitting and improving prediction accuracy, this ridge type penalty is also very computationally efficient. 

Random forest \cite{breiman2001random} is an ensemble learning method which averages over a large collection of de-correlated decision trees. Similar as bagging, random forest builds decision tress on bootstrapped samples. But unlike bagging method, when building the decision tree, each time random forest only use a portion of randomly selected features. Thus, it decorrelates the decision thees and makes their ensemble less variable. Random forest allows for interaction effects among features just like any tree based algorithm, but it corrects for the likely overfitting of decision trees. The performance of random forest is comparable to boosting, and they are easier to train and tune \cite{friedman2001elements}.

Gradient boosting machine (GBM) \cite{friedman2001greedy,friedman2002stochastic} is an ensemble method which combines weak classifiers to form a strong classifier. We use decision tree as our weak classifier. Unlike random forest which fit a large number of decision trees in parallel, GBM works in a forward ``stagewise'' fashion. In each step, GBM firstly calculate the pseudo-residuals (negative first-order derivative of the loss function) at the current model, and then fit a decision tree to the pseudo-residuals. GBM then add the fitted decision tree to the previous model. There are many parameters that we can tune in GBM. For example, we can set the order of interaction we want to consider by specifying the depth of the decision tree; We can avoid overfitting by specifying a small learning rate in GBM. GBM is also very flexible as users can provide their own loss function. GBM has been implemented in many data mining competition winning strategies. 

\subsection{Evaluation}

Evaluation is the key step to obtain feedback and find the approach that works well predicting faults with the data at hand. In this competition, each team was allowed to submit a set of predictions only once a week to score their model. However, this is not frequent enough given that there are a large number of possible models and tuning parameters to try. To remedy this problem and allow us to try many approaches, we built our own evaluation system, based on the idea of cross validation.  Our cross validation system was basically designed to mimic the competition evaluation/scoring system.

To build our own scoring system, we randomly remove 50\% of the faults in the second half of each training dataset. We build the model using the remaining fault data and attempt to predict the deleted events. We then compare the predicted faults $E_P$ with the deleted true events $E_T$, and score our model. If a fault event in $E_T$ has been correctly predicted in $E_P$ (i.e., there exists an event in $E_P$ with start time and end time within one hour of actual start time and end time, and fault type also matches), it is a true positive and receives 10 point. If a fault event in $E_P$ has correct start time and end time, and incorrect fault type, it is a misclassification and that prediction receives -0.01 point. If a fault event in $E_P$ has incorrect start time or end time, it is considered as a false positive and receives -0.1 point. If a fault event in $E_T$ has not been identified in $E_P$, it is considered a false negative and receives -0.1 point.  

We found that the above scoring system worked very well in the sense that the order of magnitude of improvement of one classification algorithm over another based on our scoring system was similar to the improvement seen on the leader board. In this way, we could use our scoring system and experiment with many different algorithms and tuning parameters. The final model achieves a score of 79570 on the training data, and 21015 on the validation data. The average score per plant of the final model on the training data is 2411 with 90\% confidence interval from 60 to 5006. The average score per plant on the validation data is 1401.

%
%

\subsection{Feature engineering}
We did feature engineering and added features like month, hour, weekday and time (the number of minutes since 00:00 of the first day of the corresponding year / (60*24)) in our classification models to predict faults' start and end time. A complete list of the features to predict faults' start and end time is given in Table \ref{all_features}. Here elapsed\_t represents elapsed time since the fault first occurred which is defined in Section 3.5. 

We also added lagged covariates of all sensor readings (R1-R4, S1-S4, E2 and E3) to the model. Specifically, for any sensor reading, $X(t)$, we included $X(t-k)$, for all nonzero $k$, where $k\geq \textrm{min\_lag}$ and $k\leq \textrm{max\_lag}$. To define these new variables we introduce the following notation: $Lk\_mi\_Sj(t) = mi\_Sj(t-k)$, where $k > 0$ and thus, represents the lagged covariate of $mi\_Sj$. In contrast, $Rk^*\_mi\_Sj(t) = mi\_Sj(t-k)$, where $k < 0$ and $k^*=-k$, and thus, represents the lead (future) covariate of $mi\_Sj$. The time interval is 15 minutes. Based on the description of the competition, we know a fault is independent of data outside a three hour window of time. So the smallest \textrm{min\_lag} and the largest \textrm{max\_lag} we considered are -12 and 12, respectively. All covariates were standardized to have mean 0 and variance 1 before feeding to the classifiers.

\begin{table*}[ht]\small 
	\begin{center}  
		\begin{tabular}{ll}
			\toprule
			\textbf{model} & \textbf{features in the model}\\
			\midrule
			predict start time &  \begin{tabular}{@{}c@{}} month, hour, weekday, time, R1-R4, S1-S4, E2, E3\\ and lagged covariates of all sensor readings\end{tabular}\\
			predict end time &  \begin{tabular}{@{}c@{}} month, hour, weekday, time, R1-R4, S1-S4, E2, E3, elapsed\_t\\ and lagged covariates of all sensor readings\end{tabular}\\
			\bottomrule
		\end{tabular}
		\caption{Features include in the models.}
		\label{all_features}
	\end{center}
\end{table*}  

\subsection{Predict Start Time}
For every plant and fault type in the cross validation training,  test or validation datasets, we built a separate classification model to predict the start time of deleted events. $Start\_Fk$, the binary indicator of whether a type $k$ fault starts within one hour, is the response variable $Y$. To train each model, we include all data from the first half of the sample where we know exactly when all faults do or do not occur. In addition, we also include data from the second half where $start\_Fk(t)=1$. That is, we only include the data for the faults that we know occur (i.e., the faults that have not been randomly deleted). We define  $X_{\textrm{train}}$ and $Y_{\textrm{train}}$ to be the resulting covariate and response matrices used to train the model. We stack the data from the second half that are not used to train the model to define $X_{\textrm{test}}$. $X_{\textrm{test}}$ contains the data where the response $start\_Fk$ ($Y_{test}$) is unknown. We then estimate $\hat{p}_{\textrm{test}}$ $(p_{\textrm{test}}=\Pr({Y}_{\textrm{test}}=1))$, and predict deleted fault start time based on the magnitude of $\hat{p}_{\textrm{test}}$. We specify our model as follows to gain the optimal performance. The optimal tuning parameter and thresholds are found by cross validation.

\begin{figure}[b]
	\centering
	\includegraphics[scale=0.4]{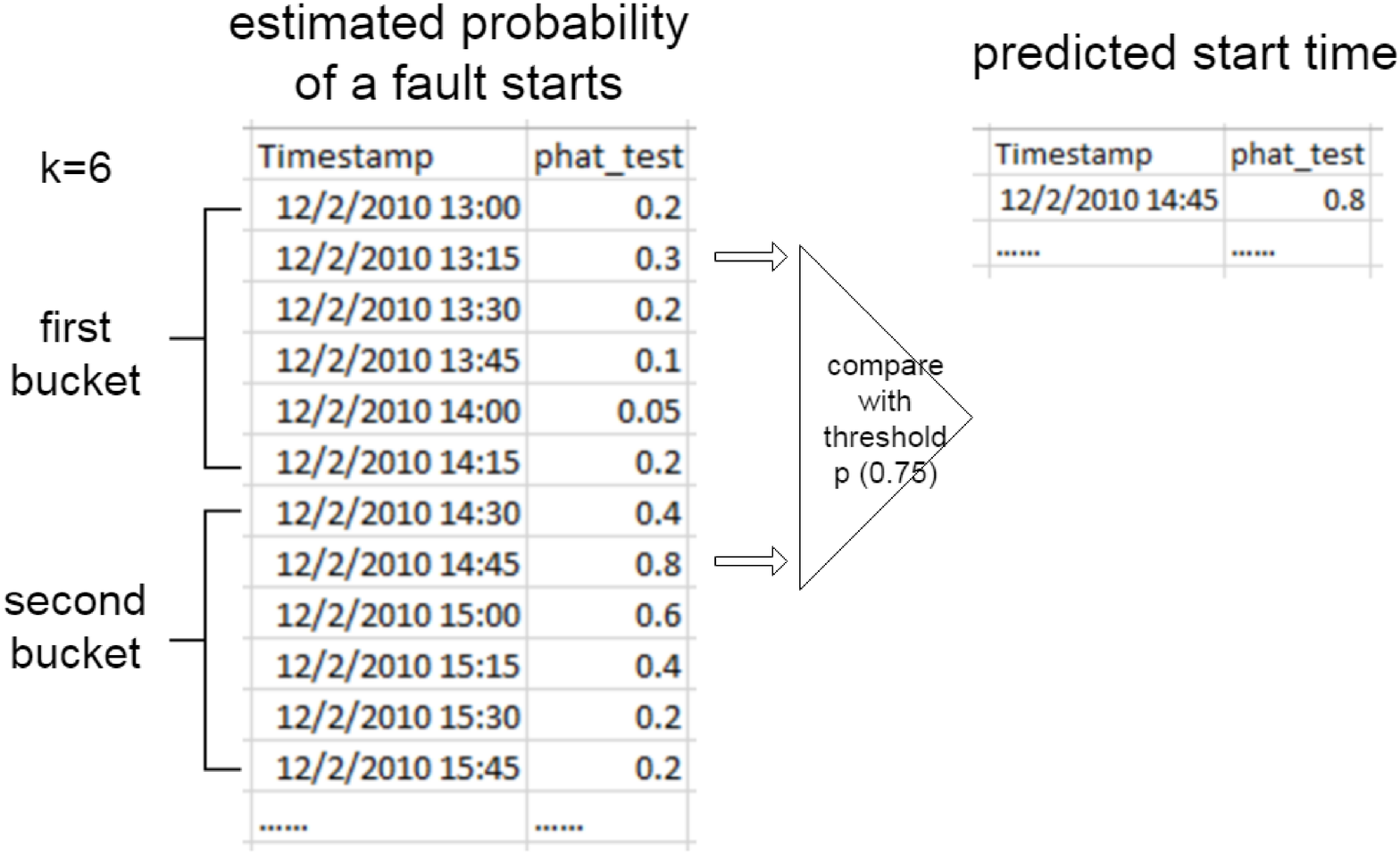}
	\caption{An example to predict the fault's start time.}
	\label{fig:predict_start_time}
\end{figure}

\begin{itemize}
	\item We tried different $\textrm{min\_lag}$ and $\textrm{max\_lag}$ combinations, and the best one we found is $\textrm{min\_lag}=-8$ and $\textrm{max\_lag}=4$. 
	\item For $k$ consecutive estimates of $p_{\textrm{test}}$ (i.e., the estimated probability a fault starts for $k$ consecutive time intervals), we found the largest probability and compared it to a threshold $p$. If it exceeded $p$, the corresponding timestamp was saved as a predicted start time. We tested different combinations of values for $k$ and $p$. The best performing combination was $k=6$ and $p=0.75$. See Figure~\ref{fig:predict_start_time} for an illustration. 
	\item We compared the performance of the various algorithms modeling covariates (month, hour, S3, R1, R2, R3, R4, etc) as categorical versus continuous variables. No real improvement was made modeling them as categorical variables, so all covariates are treated as continuous. 
	\item We experimented with various different classifiers including KNN, Naive Bayes, GBM, random forest, and penalized logistic regression. We found that random forest and penalized logistic regression performed the best. Our final algorithm was an ensemble of these latter two models, where we kept all predicted start times from random forest, and then added all predicted start times that were found in penalized logistic regression but not found in random forest.   
\end{itemize}  

One can determine which covariates are most important in predicting fault start times by looking at the random forest results. Each covariate can be scored based on mean decrease in impurity. Specifically, we add up the total amount that the Gini index is decreased by splits over a given predictor averaged over all trees, and this value is the measurement of feature importance. Gini index is defined as
$$ G = \sum_{i=1}^{K}\hat{p}_{mk}(1-\hat{p}_{mk}),$$
where $\hat{p}_{mk}$ represents the proportion of observations in the $m$th node that belongs to the $k$th class and $K$ is the total number of classes \cite{james2013introduction}. Table \ref{import_cov_start} lists the top 15 most important covariates and their corresponding score (standardized) for each one of the five fault types in plant 1 and 6. The importance of covariates vary from fault to fault and from plant to plant. S3 is the most important covariate in predicting the start time of F1 in plant 1, while both R1 and S3 seem to be important in predicting F1's start time in plant 6; S3 is the most important covariate in predicting F2's start time, while E3 is the most important covariate in predicting F4's start time in plant 1. 

Table \ref{import_cov_start2} shows the percentage of times that each covariate ranked in the top 15 importance score (random forest to predict fault start time) averaged over all plants. S1-S4 seem to be more important than R1-R4, and the covariates time, month, E2 and E3 are also important features in the models.

\begin{table*}[ht]\footnotesize 
	\setlength{\tabcolsep}{2pt}   
	\begin{center}  
		\caption{Top 15 most important covariates to predict fault start time for each of five faults in plant 1 and 6.}
		\label{import_cov_start}
		\begin{tabular}{rrlrlrlrlrlr}
			\toprule
			\textbf{Plant} & \textbf{Rank} & \multicolumn{2}{c}{\textbf{F1}} & \multicolumn{2}{c}{\textbf{F2}} & \multicolumn{2}{c}{\textbf{F3}} & \multicolumn{2}{c}{\textbf{F4}} & \multicolumn{2}{c}{\textbf{F5}} \\
			\cmidrule(r){3-4} \cmidrule(lr){5-6} \cmidrule(lr){7-8} \cmidrule(lr){9-10} \cmidrule(l){11-12}
			&  & covariate &  score & covariate &  score & covariate &  score & covariate &  score & covariate &  score \\
			\midrule
			1 &     1 &     R4\_m4\_S3 &    0.0136 &     R1\_m5\_S3 &    0.0293 &     R7\_m2\_S4 &    0.0180 &     R5\_n3\_E3 &    0.0241 &     R4\_m3\_R1 &    0.0131 \\
			&     2 &     R5\_m4\_S3 &    0.0118 &     R4\_m5\_S3 &    0.0268 &     R7\_m2\_S2 &    0.0179 &        n3\_E3 &    0.0230 &     R8\_m1\_S1 &    0.0105 \\
			&     3 &     R3\_m4\_S3 &    0.0104 &     R8\_m5\_S3 &    0.0247 &     R6\_m2\_S4 &    0.0163 &     R1\_n3\_E3 &    0.0194 &     R5\_m5\_S4 &    0.0099 \\
			&     4 &         time &    0.0095 &     R3\_m5\_S3 &    0.0245 &     R1\_m5\_S3 &    0.0153 &     R3\_n3\_E3 &    0.0179 &     R3\_m3\_R1 &    0.0091 \\
			&     5 &     R8\_m6\_S3 &    0.0089 &     R2\_m5\_S3 &    0.0241 &     R2\_m5\_S3 &    0.0147 &     R6\_n3\_E3 &    0.0176 &     R7\_m5\_S2 &    0.0085 \\
			&     6 &     R2\_m4\_S3 &    0.0088 &     R6\_m5\_S3 &    0.0229 &     R6\_m5\_S3 &    0.0146 &     R8\_n3\_E3 &    0.0166 &     R5\_m3\_R1 &    0.0084 \\
			&     7 &     R7\_m4\_S3 &    0.0086 &     R7\_m5\_S3 &    0.0190 &     R6\_m2\_S2 &    0.0145 &     R2\_n3\_E3 &    0.0160 &     R7\_m1\_S1 &    0.0082 \\
			&     8 &     L4\_m6\_S3 &    0.0073 &     R5\_m5\_S3 &    0.0181 &     R1\_m2\_S2 &    0.0143 &     R4\_n3\_E3 &    0.0139 &     R4\_m5\_S2 &    0.0081 \\
			&     9 &        month &    0.0072 &     L1\_m5\_S3 &    0.0169 &     R8\_m2\_S4 &    0.0140 &     R7\_n3\_E3 &    0.0137 &     R3\_m1\_S3 &    0.0080 \\
			&    10 &     R5\_m6\_S3 &    0.0067 &        m5\_S3 &    0.0119 &     R3\_m5\_S3 &    0.0137 &     R1\_m2\_S2 &    0.0131 &        m6\_R1 &    0.0080 \\
			&    11 &     R6\_m4\_S3 &    0.0067 &     L4\_m5\_S3 &    0.0108 &        m5\_S3 &    0.0133 &     L1\_m2\_S2 &    0.0124 &     R1\_m1\_R1 &    0.0076 \\
			&    12 &     R1\_m4\_S3 &    0.0060 &     R7\_m2\_S2 &    0.0095 &     R4\_m5\_S3 &    0.0130 &     L2\_n3\_E3 &    0.0116 &     R5\_m1\_S3 &    0.0076 \\
			&    13 &        m4\_S3 &    0.0059 &     R8\_m2\_S4 &    0.0088 &     R4\_m2\_S2 &    0.0125 &     R1\_m2\_S4 &    0.0114 &     L4\_m2\_S3 &    0.0073 \\
			&    14 &     R4\_m5\_S3 &    0.0059 &     R7\_m2\_S4 &    0.0086 &     R8\_m2\_S2 &    0.0123 &     L1\_n3\_E3 &    0.0110 &     L1\_m2\_S3 &    0.0072 \\
			&    15 &     R6\_m6\_S3 &    0.0058 &     R5\_m2\_S3 &    0.0080 &     R5\_m2\_S2 &    0.0119 &     L3\_m2\_S2 &    0.0104 &     R6\_m2\_S3 &    0.0070 \\
			\midrule
			6 &     1 &     R7\_m1\_S3 &    0.0110 &     L4\_m7\_S1 &    0.0116 &     L4\_m7\_S1 &    0.0151 &     R5\_m8\_S1 &    0.0187 &     R5\_m8\_S4 &    0.0244 \\
			&     2 &     R3\_m4\_R1 &    0.0103 &     R3\_m9\_S3 &    0.0095 &     R4\_m9\_R2 &    0.0130 &     R6\_m8\_S1 &    0.0170 &     R4\_n2\_E3 &    0.0237 \\
			&     3 &     R1\_m4\_R1 &    0.0095 &     R2\_m9\_S3 &    0.0090 &     R4\_m9\_S3 &    0.0122 &     R4\_m8\_S1 &    0.0155 &     R4\_n1\_E3 &    0.0197 \\
			&     4 &        m4\_R1 &    0.0095 &     R1\_m9\_S3 &    0.0084 &     L4\_m3\_S1 &    0.0119 &     R8\_m8\_S1 &    0.0151 &     R4\_m8\_S2 &    0.0195 \\
			&     5 &     L2\_m4\_R1 &    0.0091 &     R4\_m9\_S3 &    0.0075 &         hour &    0.0110 &     R7\_m8\_S1 &    0.0138 &         time &    0.0160 \\
			&     6 &     R7\_m4\_R1 &    0.0090 &     L3\_m7\_S1 &    0.0074 &     R8\_m9\_S1 &    0.0100 &     R2\_m8\_S1 &    0.0137 &     R1\_m8\_S2 &    0.0147 \\
			&     7 &     R5\_m4\_R1 &    0.0079 &     R4\_n1\_E3 &    0.0070 &     R3\_m9\_S3 &    0.0092 &     R1\_m8\_S1 &    0.0124 &     R1\_m8\_S4 &    0.0147 \\
			&     8 &     R8\_m1\_S3 &    0.0078 &     L1\_m7\_S1 &    0.0064 &     R4\_m9\_R3 &    0.0087 &     R3\_m8\_S1 &    0.0113 &     R5\_m4\_R1 &    0.0146 \\
			&     9 &     R7\_m7\_S3 &    0.0072 &     R7\_m7\_S3 &    0.0064 &     R4\_m9\_R4 &    0.0084 &        m8\_S1 &    0.0092 &     R7\_m4\_R1 &    0.0141 \\
			&    10 &     R4\_m7\_S3 &    0.0070 &        m7\_S1 &    0.0064 &    R5\_m10\_R2 &    0.0082 &     R5\_n1\_E3 &    0.0069 &     R3\_m4\_R3 &    0.0140 \\
			&    11 &         time &    0.0069 &         time &    0.0062 &     R4\_m9\_S1 &    0.0082 &     R4\_n1\_E3 &    0.0064 &     R3\_m4\_R1 &    0.0132 \\
			&    12 &     R2\_m7\_S3 &    0.0061 &     L4\_m3\_S1 &    0.0061 &     R6\_m9\_S1 &    0.0079 &     R2\_n2\_E3 &    0.0061 &     R5\_n1\_E3 &    0.0129 \\
			&    13 &     R4\_m4\_R1 &    0.0061 &     L2\_m7\_S1 &    0.0059 &     R3\_m9\_S1 &    0.0078 &     R3\_n2\_E3 &    0.0059 &     R7\_n2\_E3 &    0.0128 \\
			&    14 &     R7\_n1\_E2 &    0.0060 &    R5\_m10\_R2 &    0.0056 &     R5\_m9\_S1 &    0.0078 &     R2\_n1\_E3 &    0.0059 &     R4\_m4\_R3 &    0.0127 \\
			&    15 &     R6\_m7\_S3 &    0.0059 &     R5\_m9\_S3 &    0.0053 &    R5\_m10\_R4 &    0.0076 &     R5\_n2\_E3 &    0.0057 &     R5\_n2\_E3 &    0.0126 \\
			\bottomrule
		\end{tabular}
	\end{center}
\end{table*}

\begin{table}[ht]   
	\begin{center}  
		\caption{Percentage that each covariate rank top 15 in the importance score (random forest to predict fault start time) averaged over all plants.}
		\label{import_cov_start2}
		\begin{tabular}{lrrrrr}
			\toprule
			\textbf{Covariate} &  \textbf{F1} &  \textbf{F2} &  \textbf{F3} &  \textbf{F4} &  \textbf{F5} \\
			\midrule
			time    & 60\% & 48\% & 31\% & 32\% & 18\% \\
			month   & 27\% & 14\% & 17\% & 11\% &  4\% \\
			hour    &  5\% & 12\% &  8\% &  4\% &  0\% \\
			weekday &  2\% &  0\% &  0\% &  2\% &  2\% \\
			S1      & 32\% & 40\% & 81\% & 82\% & 64\% \\
			S2      & 54\% & 48\% & 58\% & 36\% & 88\% \\
			S3      & 65\% & 78\% & 33\% & 34\% & 20\% \\
			S4      & 49\% & 52\% & 56\% & 41\% & 90\% \\
			R1      & 19\% &  5\% & 10\% & 14\% & 10\% \\
			R2      & 17\% & 19\% & 23\% & 16\% & 14\% \\
			R3      & 19\% & 12\% &  8\% & 12\% & 12\% \\
			R4      &  2\% &  3\% &  6\% &  4\% &  4\% \\
			E2      & 19\% & 31\% & 12\% & 25\% &  8\% \\
			E3      & 25\% & 41\% & 17\% & 39\% & 16\% \\
			\bottomrule
		\end{tabular}
	\end{center}
\end{table}   

\subsection{Predict End Time}

To predict the end time of a plant fault, we built another classification model. As with the start time prediction problem, we estimate a separate model for each fault type and plant. To explain our modeling approach, suppose we want to find the end time of a predicted type 1 fault ($F1$). We first estimate an upper bound for the duration of a $F1$ fault ($t_{\textrm{max}}$) based on all known $F1$ events. We estimate $t_{\textrm{max}}$ as follows: $t_{\textrm{max}}=\textrm{max}(8, q_{0.95})$, where $q_{0.95}$ is the 95\% upper quantile of all historical F1 durations. 

Intuitively, we could predict the fault end time by calculating the elapsed time since the fault first occurred. We denote it as elapsed\_t which is measured in units of 15 minutes. We find that the model only based on elapsed\_t isn't accurate enough, so we add other covariates to the model. 

The classification model is trained using data from the $t_{\textrm{max}}$ intervals following the onset of each known $F1$ event. These data, stacked together, form the matrix of covariates for the training model ($X_{\textrm{train}}$). $end\_F1$, the binary indicator of whether fault 1 ends within one hour of the corresponding timestamp, serves as the response variable, $Y_{\textrm{train}}$. Once the classification model is trained, it is used to estimate the probability that each of our predicted events will end in any one of the $t_{\textrm{max}}$ time periods following the predicted event start time. These predictions are based on $X_{\textrm{test}}$, formed by stacking the $t_{\textrm{max}}$ intervals following the onset of each predicted $F1$ event.   

\begin{figure}[ht]
	\centering
	\includegraphics[scale=0.2]{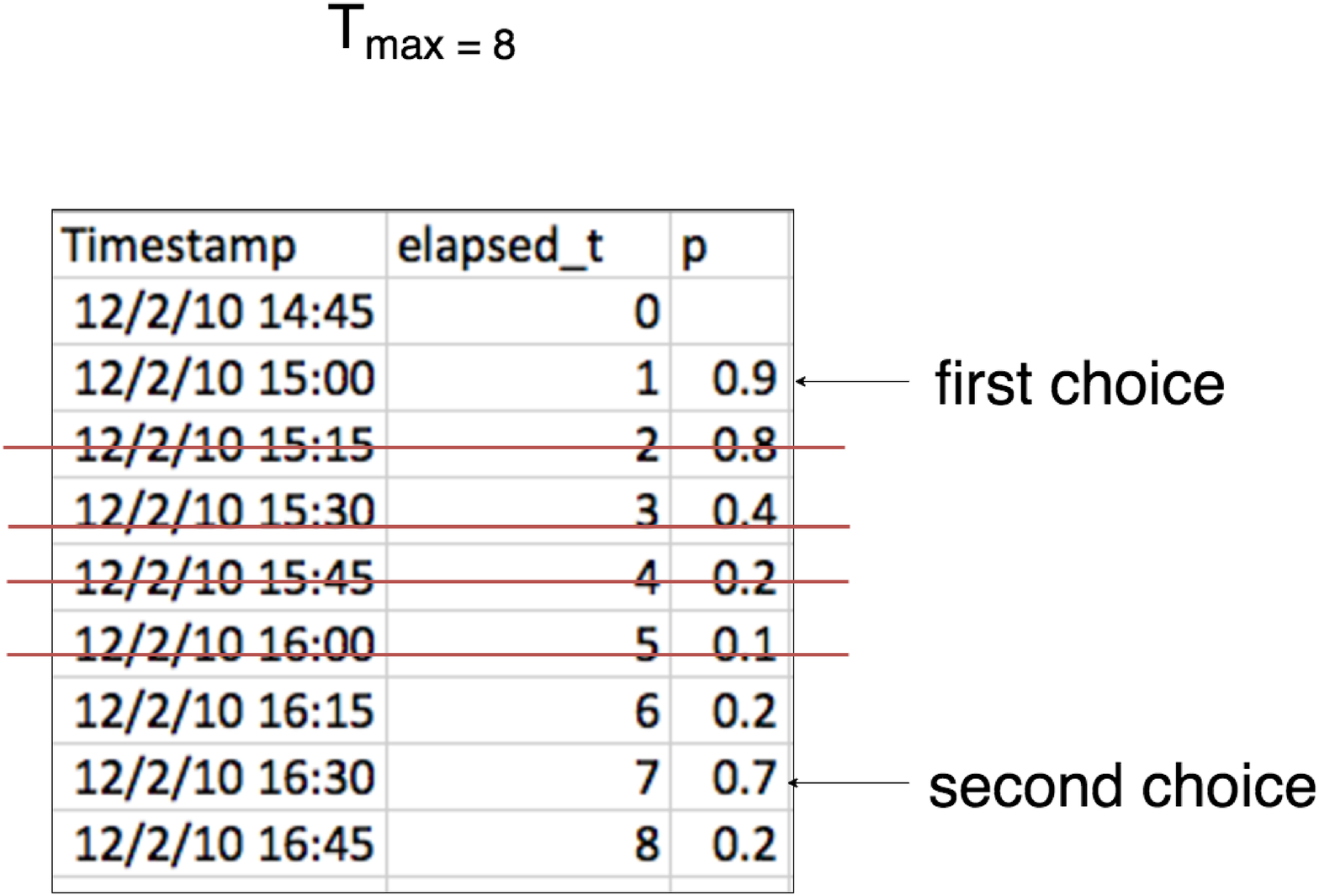}
	\caption{An example to predict the fault's end time.}
	\label{fig:predict_end_time}
\end{figure}

Given the small penalty for false negative predictions relative to the reward for a true positive prediction, we allow our system to predict as many as two end times for each predicted event. The first estimated end time is made by finding the time period within the $t_{\textrm{max}}$ periods following our predicted event with the largest estimated probability that $end\_F1$=1. This is our first end time prediction. To look for a possible second prediction, we delete all observations with timestamps within one hour of our first estimated end time. We then find the (remaining) time period with the largest probability. If the estimated probability in this period is larger than our threshold $p2$, this is a second end time prediction. If the probability is less than $p2$, only one end time is predicted. See Figure~\ref{fig:predict_end_time} for an illustration, where the second end time prediction is $elapsed\_t=7$ and is kept as $\hat{p}>p2(=0.2)$.

The following list details the specifics of our final algorithm for the end time classification problem. The optimal tuning parameter and thresholds are found by cross validation.

\begin{itemize}
	\item The optimal threshold probability for deciding on a second end time is $p2$=0.2.
	\item The optimal lag choice is $\textrm{min\_lag}=-8$ and $\textrm{max\_lag}=8$.
	\item We model all covariates as continuous variables. 
	\item We have compared the performance of various different classifier methodologies including GBM, random forest, and penalized logistic regression. We find that GBM has the best performance. We choose tree\_number=200 and tree\_depth=5 for the GBM.
\end{itemize}  

As with the start times, we find the most important covariates in predicting fault end time by calculating a score based on mean decrease impurity in GBM. In Table \ref{import_cov_end} we list the top 15 most important covariates and their corresponding score in predicting fault end time for each one of the five fault types in plant 1 and 6. The most important covariate is elapsed\_t, which ranks number one in all cases. 

Table \ref{import_cov_end2} shows the percentage of times that each covariate ranked in the top 15 importance score (GBM to predict fault end time) averaged over all plants. elapsed\_t is the most important covariate. S1-S4 are again more important than R1-R4, and the covariates time, hour, weekday, E2 and E3 are also important features in the models.  

\begin{table*}[ht]\footnotesize
	\setlength{\tabcolsep}{2pt}  
	\begin{center}  
		\caption{Top 15 most important covariates to predict fault end time for each of five faults in plant 1 and 6.}
		\label{import_cov_end}
		\begin{tabular}{rrlrlrlrlrlr}
			\toprule
			\textbf{Plant} & \textbf{Rank} & \multicolumn{2}{c}{\textbf{F1}} & \multicolumn{2}{c}{\textbf{F2}} & \multicolumn{2}{c}{\textbf{F3}} & \multicolumn{2}{c}{\textbf{F4}} & \multicolumn{2}{c}{\textbf{F5}} \\
			\cmidrule(r){3-4} \cmidrule(lr){5-6} \cmidrule(lr){7-8} \cmidrule(lr){9-10} \cmidrule(l){11-12}
			&  & covariate &  score & covariate &  score & covariate &  score & covariate &  score & covariate &  score \\
			\midrule
			1 &     1 &           elapsed\_t &    0.0707 &           elapsed\_t &    0.1406 &           elapsed\_t &    0.2616 &           elapsed\_t &    0.2444 &           elapsed\_t &    0.1472 \\
			&     2 &         time &    0.0285 &         time &    0.0338 &         time &    0.0374 &     L1\_m3\_S1 &    0.0186 &     L1\_m3\_S1 &    0.0678 \\
			&     3 &     L1\_m4\_S3 &    0.0182 &     L1\_m5\_S1 &    0.0121 &     L1\_m5\_S1 &    0.0101 &     L2\_m3\_S1 &    0.0131 &         time &    0.0574 \\
			&     4 &     R8\_m4\_S3 &    0.0155 &     R8\_m2\_S3 &    0.0116 &         hour &    0.0098 &     L3\_m6\_S1 &    0.0127 &     L2\_m3\_S1 &    0.0516 \\
			&     5 &     R7\_m4\_S3 &    0.0110 &     R8\_m2\_S1 &    0.0110 &     L1\_m2\_S1 &    0.0092 &     L2\_m6\_R1 &    0.0104 &     L3\_m2\_S1 &    0.0159 \\
			&     6 &     L2\_m4\_S3 &    0.0093 &     L1\_m2\_S1 &    0.0095 &     L8\_m2\_S1 &    0.0076 &     R6\_n3\_E2 &    0.0088 &     L3\_m3\_S1 &    0.0142 \\
			&     7 &     R8\_m1\_S4 &    0.0076 &     R7\_m2\_S1 &    0.0091 &     R3\_n2\_E2 &    0.0074 &     R4\_m3\_S1 &    0.0085 &     L6\_m5\_S1 &    0.0093 \\
			&     8 &     L1\_m4\_S4 &    0.0070 &         hour &    0.0083 &      weekday &    0.0065 &         time &    0.0083 &     L1\_m3\_S2 &    0.0088 \\
			&     9 &     R8\_m1\_S3 &    0.0069 &     R6\_m2\_S1 &    0.0079 &     R8\_m4\_S4 &    0.0059 &     R2\_m3\_S1 &    0.0075 &     R8\_m1\_S1 &    0.0087 \\
			&    10 &     R5\_m4\_S3 &    0.0067 &     L1\_m3\_S1 &    0.0075 &     R8\_m2\_S4 &    0.0056 &     R6\_n3\_E3 &    0.0075 &     R8\_m3\_S1 &    0.0075 \\
			&    11 &     R6\_m4\_S4 &    0.0066 &     L2\_m5\_S1 &    0.0073 &     L2\_m5\_S1 &    0.0055 &     R5\_m2\_S1 &    0.0074 &     R8\_m1\_S4 &    0.0074 \\
			&    12 &     R7\_m1\_S4 &    0.0064 &     R8\_m5\_S1 &    0.0067 &     L2\_m2\_S1 &    0.0051 &      weekday &    0.0073 &     L1\_m2\_S1 &    0.0068 \\
			&    13 &     R8\_m4\_S4 &    0.0063 &     R8\_m5\_S4 &    0.0063 &     L5\_m2\_S1 &    0.0048 &     L1\_m6\_R1 &    0.0068 &     L8\_m6\_S1 &    0.0066 \\
			&    14 &     L2\_m1\_S4 &    0.0062 &     L1\_m2\_S4 &    0.0060 &     L8\_m1\_S1 &    0.0048 &     R8\_n3\_E3 &    0.0066 &     L5\_m4\_S1 &    0.0064 \\
			&    15 &     L2\_m4\_S4 &    0.0061 &     R8\_m5\_S2 &    0.0060 &     R7\_n3\_E2 &    0.0046 &     R3\_n3\_E2 &    0.0063 &     L1\_m3\_S4 &    0.0064 \\
			\midrule
			6 &     1 &           elapsed\_t &    0.1210 &           elapsed\_t &    0.1633 &           elapsed\_t &    0.0934 &           elapsed\_t &    0.3109 &           elapsed\_t &    0.1330 \\
			&     2 &         time &    0.0292 &         time &    0.0434 &     L1\_m9\_S1 &    0.0683 &     L2\_m8\_S1 &    0.0380 &         time &    0.0276 \\
			&     3 &     R8\_m9\_S3 &    0.0111 &     R8\_m9\_S3 &    0.0144 &     R8\_m9\_S1 &    0.0365 &     L1\_m8\_S1 &    0.0366 &     R8\_m8\_S4 &    0.0151 \\
			&     4 &     L1\_m9\_S3 &    0.0100 &     L1\_m9\_S3 &    0.0114 &        m6\_S1 &    0.0229 &     L8\_m3\_S4 &    0.0263 &     L1\_m8\_S2 &    0.0147 \\
			&     5 &    R8\_m10\_S3 &    0.0081 &     L1\_m9\_S1 &    0.0067 &     L1\_m6\_S1 &    0.0224 &     L8\_m3\_S2 &    0.0167 &     R8\_m8\_S2 &    0.0144 \\
			&     6 &         hour &    0.0071 &     L1\_m9\_S2 &    0.0060 &     L1\_m3\_S1 &    0.0162 &     L1\_m3\_S4 &    0.0166 &     L1\_m8\_S4 &    0.0118 \\
			&     7 &    L1\_m10\_S3 &    0.0059 &     R7\_m9\_S3 &    0.0059 &     L1\_m4\_R1 &    0.0161 &     L1\_m6\_S1 &    0.0164 &     R8\_m8\_S1 &    0.0113 \\
			&     8 &    L1\_m10\_S4 &    0.0058 &     L1\_m9\_S4 &    0.0055 &     L6\_m9\_S4 &    0.0115 &     L1\_m3\_S2 &    0.0151 &     R6\_m8\_S1 &    0.0081 \\
			&     9 &     L1\_m7\_S2 &    0.0052 &     R8\_m1\_S3 &    0.0053 &        n1\_E2 &    0.0095 &        m7\_S1 &    0.0124 &     L1\_m8\_S1 &    0.0073 \\
			&    10 &     R8\_m9\_S4 &    0.0051 &     L2\_m9\_S3 &    0.0048 &        m8\_S1 &    0.0086 &     R8\_m8\_S1 &    0.0117 &     L2\_m8\_S1 &    0.0072 \\
			&    11 &     L1\_m7\_S4 &    0.0046 &     R8\_n2\_E2 &    0.0047 &     L6\_m9\_S2 &    0.0082 &     L1\_m7\_S1 &    0.0109 &     L8\_m8\_S1 &    0.0072 \\
			&    12 &      weekday &    0.0045 &     R8\_m9\_S1 &    0.0047 &     L2\_m9\_S1 &    0.0076 &     R8\_n1\_E3 &    0.0092 &     R4\_m8\_S1 &    0.0070 \\
			&    13 &    R7\_m10\_S1 &    0.0044 &     L1\_m1\_S3 &    0.0044 &     L1\_m2\_S1 &    0.0075 &        m3\_S2 &    0.0089 &     L7\_m2\_S1 &    0.0069 \\
			&    14 &     R8\_n1\_E3 &    0.0043 &         hour &    0.0042 &     R5\_m9\_S4 &    0.0059 &    L1\_m10\_S1 &    0.0085 &     R8\_m1\_S1 &    0.0067 \\
			&    15 &     R8\_m9\_S1 &    0.0042 &     L1\_m7\_S3 &    0.0042 &     R7\_m4\_S4 &    0.0057 &     R4\_m6\_S1 &    0.0081 &     R5\_m8\_S1 &    0.0062 \\
			\bottomrule
		\end{tabular}
	\end{center}
\end{table*}   

\begin{table}[ht]\small   
	\begin{center}  
		\caption{Percentage that each covariate rank top 15 in the importance score (GBM to predict fault end time) averaged over all plants.}
		\label{import_cov_end2}
		\begin{tabular}{lrrrrr}
			\toprule
			\textbf{Covariate} &  \textbf{F1} &  \textbf{F2} &  \textbf{F3} &   \textbf{F4} &   \textbf{F5} \\
			\midrule
			elapsed\_t & 98\% & 95\% & 93\% &  93\% &  95\% \\
			time      & 94\% & 89\% & 78\% &  79\% &  63\% \\
			month     &  0\% &  0\% &  0\% &   0\% &   0\% \\
			hour      & 57\% & 58\% & 50\% &  45\% &  12\% \\
			weekday   & 21\% & 25\% & 35\% &  30\% &  12\% \\
			S1        & 65\% & 86\% & 98\% & 100\% & 100\% \\
			S2        & 90\% & 91\% & 91\% &  73\% &  98\% \\
			S3        & 89\% & 82\% & 20\% &  12\% &  14\% \\
			S4        & 97\% & 93\% & 93\% &  73\% &  95\% \\
			R1        & 11\% & 18\% & 15\% &  25\% &   9\% \\
			R2        &  2\% &  2\% &  4\% &   7\% &   5\% \\
			R3        &  2\% &  7\% & 11\% &   9\% &   7\% \\
			R4        &  0\% &  2\% &  7\% &   7\% &   7\% \\
			E2        & 22\% & 33\% & 35\% &  30\% &  16\% \\
			E3        & 29\% & 37\% & 37\% &  32\% &  40\% \\
			\bottomrule
		\end{tabular}
	\end{center}
\end{table}  

\section{Conclusion}  
In this paper, we proposed and implemented a machine learning based algorithm to detect industrial plant faults. The encouraging results demonstrated the usefulness of data-driven algorithms in fault detection of complex systems. Several extensions to our algorithms were considered but not implemented due to the time constraints of the PHM Society Data Challenge.  These additional approaches are left as future work. 

One such approach would be to not model each plant independently.  Alternatively, we could try to first group the plants into clusters of like plants (based on like distributions and/or timing of faults, for example), and then model plants in each cluster together.  

Another untried approach is deep learning neural networks. Convolutional neural networks or recurrent neural networks, which have been shown to be powerful tools when modeling with large and complex datasets, may yield good results. Convolutional neural network can automatically consider lagged observations by modeling temporal contiguous observations jointly together. Recurrent neural network can create an internal state of the network which allows it to exhibit dynamic temporal behavior. These facts make deep learning neural networks potentially very useful in fault detection for the PHM data.

Lastly,  in our approach, lagged covariates are added to the model which creates high dimensional features. Curse of dimensionality may damnify the classifiers' performances. Techniques such as principal component analysis and functional data analysis can be applied to extract key features from time series covariates and reduce the feature dimension.  

\section*{Acknowledgment}
The author wants to give thanks to SAS colleges Anya Mcguirk, Sergiy Peredriy, Arin Chaudhuri, Alex Chien, Deovrat Kakde and Gul Ege for their help in the Prognostics and Health Management Society 2015 data challenge competition.

\bibliographystyle{apalike}
\bibliography{ijphm}

\end{document}